\documentclass{article}
\usepackage{bbm}
\usepackage{amsmath}
\usepackage{amsthm}
\usepackage{epsfig}
\usepackage{psfrag}

\newcommand{\F}{\ensuremath{\mathcal F}}
\renewcommand{\H}{\ensuremath{\mathcal H}}
\newcommand{\pot}{\ensuremath{\mathcal P}}
\newcommand{\X}{\ensuremath{\mathcal X}}
\newcommand{\Y}{\ensuremath{\mathcal Y}}
\newcommand{\NNN}{\mathbbm{N}}
\newcommand{\m}{d}
\newcommand{\clup}{c.u.p.}

\def\squareforqed{\hbox{\rlap{$\sqcap$}$\sqcup$}}
\def\qed{\ifmmode\squareforqed\else{\unskip\nobreak\hfil
\penalty50\hskip1em\null\nobreak\hfil\squareforqed
\parfillskip=0pt\finalhyphendemerits=0\endgraf}\fi}

\theoremstyle{plain}
\newtheorem{theorem}{Theorem}
\newtheorem{corollary}{Corollary}

\newtheorem{lemma}{Lemma}
\theoremstyle{remark}
\newtheorem{remark}{Remark}

\begin{document}
\title{\textbf{On Classes of Functions for which\\No Free Lunch Results Hold}}
\author{Christian Igel and Marc Toussaint\\
Institut f\"ur Neuroinformatik\\Ruhr-Universit\"at Bochum,
 Germany\\\normalsize\{Christian.Igel, Marc.Toussaint\}@neuroinformatik.ruhr-uni-bochum.de}
\date{}
\maketitle

\hyphenation{ex-plo-ra-tion}

\renewcommand{\theenumi}{\alph{enumi}}
\renewcommand{\labelenumi}{(\alph{enumi})}

\begin{abstract}
\noindent In a recent paper \cite{schuhmacher:01} it was shown that No Free
Lunch results \cite{wolpert:95} hold for any subset $F$ of the set of all
possible functions from a finite set $\X$ to a finite set $\Y$ iff $F$
is closed under permutation of \X.  In this article, we prove that the
number of those subsets can be neglected compared to the overall
number of possible subsets.  Further, we present some arguments why
problem classes relevant in practice are not likely to be closed under
permutation.
\end{abstract}

%%%%%%%%%%%%%%%%%%%%%%%%%%%%%%%%%%%%%%%%%%%%%%%%%%%%%%%%%%%%%%%%%%%%%%%%%%%
\section{Introduction}
%%%%%%%%%%%%%%%%%%%%%%%%%%%%%%%%%%%%%%%%%%%%%%%%%%%%%%%%%%%%%%%%%%%%%%%%%%%
The No Free Lunch (NFL) theorems---roughly speaking---state that all
search algorithms have the same average performance over all possible
objective functions $f:\X\to\Y$, where the search space $\X$ as well
as the cost-value space $\Y$ are finite sets \cite{wolpert:95}.
However, it has been argued that in practice one does not need an
algorithm that performs well on all possible functions, but only on a
subset that arises from the real-world problems at hand.  Further, it
has been shown that for pseudo-Boolean functions restrictions of the
complexity lead to subsets of functions on which some algorithms
perform better than others (e.g., in \cite{whitley:99} complexity is
defined in terms of the number of local minima and in \cite{droste:99}
the complexity is defined based on the size of the smallest OBDD
representations of the functions).
%In the original NFL
%the performance is  averaged over all possible objective functions.

Recently, a sharpened version of the NFL theorem has been proven that
states that NFL results hold for any subset $F$ of the set of all
possible functions if and only if $F$ is closed under permutation
\cite{schuhmacher:01}. Based on this important result, we can derive
classes of functions where NFL does not hold simply by showing that
these classes are not closed under permutation (\clup).  This leads to
the encouraging results in this paper: It is proven that the fraction
of subsets \clup{} is so small that it can be neglected. In addition,
arguments are given why we think that objective functions resulting
from important classes of real-world problems are likely not to be
\clup{}

In the following section, we give some basic definitions and
concisely restate the sharpened NFL theorem given
in \cite{schuhmacher:01}. Then we derive the number of subsets
\clup{} Finally, we discuss some observations
regarding structured search spaces and closure under permutation.

%%%%%%%%%%%%%%%%%%%%%%%%%%%%%%%%%%%%%%%%%%%%%%%%%%%%%%%%%%%%%%%%%%%%%%%%%%%
\section{Preliminaries}
%%%%%%%%%%%%%%%%%%%%%%%%%%%%%%%%%%%%%%%%%%%%%%%%%%%%%%%%%%%%%%%%%%%%%%%%%%%

We consider a finite search space $\X$ and a finite set of cost values
$\Y$.  Let $\F=\Y^\X$ be the set of all objective functions
$f:\X\to\Y$ to be optimized (also called fitness, energy, or cost
functions).  NFL theorems are concerned with non-repeating black-box
search algorithms (referred to simply as algorithms for brevity) that
choose a new exploration point in the search space depending on the
complete history of prior explorations: Let the sequence
$T_m=\left<(x_1,f(x_1)),(x_2,f(x_2)), \dots, (x_m,f(x_m))\right>$
represent $m$ non-repeating explorations $x_i \in \X$,
$\forall_{i,j}:\, x_i\not=x_j$ and their cost values $f(x_i) \in
\Y$. An algorithm $a$ appends a pair $(x_{m+1},f(x_{m+1}))$ to this
sequence by mapping $T_m$ to a new point $x_{m+1}$, $\forall_{i}:\,
x_{m+1}\not=x_i$.  Generally, the performance of an algorithm $a$
after $m$ iterations with respect to a function $f$ depends on the
sequence of cost values
$Y(f,m,a)=\left<f(x_1),f(x_2), \dots,f(x_m)\right>$
the algorithm has produced.
%to be the sequence
%of $\Y$ values 
%produced by $m$ successive applications of $a$.
Let the function $c$
denote a performance measure mapping sequences of $\Y$ to the real
numbers (e.g., in the case of function minimization a performance
measure that returns the minimum $\Y$ value in the sequence could be a
reasonable choice).

Let $\pi:\X\rightarrow\X$
be a permutation (i.e., bijective function) of $\X$.
The set of all permutations of $\X$ is denoted by $\pot(\X)$.
A set $F\subseteq\F$ is said to be closed under permutation (\clup)
if for any $\pi\in\pot(\X)$ and any function $f\in F$
the function $f\circ\pi$ is also in $F$.
%The permutation $\pi(f)$ of a function $f:\X\to\Y$
%is the function   $\pi(f):\X\to\Y$ defined 
%by $\pi(f)(x):f(\pi^{-1}(x))$.  
\begin{theorem}[NFL]\label{nfl:thm}
For any two  algorithms $a$ and $b$,
any value $k\in \mathbbm R$, and any performance measure $c$
%and any function $f\inF$
\[
\sum_{f\in F}\delta(k, c(Y(f,m,a)))
=
\sum_{f\in F}\delta(k, c(Y(f,m,b)))
\]
iff $F$ is \clup{}
\end{theorem}
\noindent Herein, $\delta$ denotes the Kronecker function ($\delta(i,j)=1$ if
$i=j$, $\delta(i,j)=0$ otherwise). A proof of theorem \ref{nfl:thm}
is given in \cite{schuhmacher:01}.
This theorem implies
%\begin{itemize}
%\item Consider 
that for any two algorithms $a$ and $b$ and any function $f_a\in F$,
where $F$ is \clup, there is a function $f_b\in F$ on which $b$ has
the same performance
as $a$ on $f_a$.
%\end{itemize}

%%%%%%%%%%%%%%%%%%%%%%%%%%%%%%%%%%%%%%%%%%%%%%%%%%%%%%%%%%%%%%%%%%%%%%%%%%%
\section{Fraction of Subsets Closed under Permutation}
%%%%%%%%%%%%%%%%%%%%%%%%%%%%%%%%%%%%%%%%%%%%%%%%%%%%%%%%%%%%%%%%%%%%%%%%%%%

Let $\F=\Y^\X$ be the set of functions mapping $\X \to \Y$.  There
exist $2^{\left(|\Y|^{|\X|}\right)}-1$ non-empty subsets of $\F$.  We want to
calculate the fraction of subsets that are \clup{}
\begin{theorem}\label{main:thm}
The number of non-empty subsets of $\Y^\X$ that are \clup{}
is given by
\begin{equation*}
2^{\binom{|\X|+|\Y|-1}{|\X|}} -1
\enspace.
\end{equation*}
\end{theorem}
\noindent The proof is given in the appendix.

\begin{figure}[b!]
\psfrag{A }[l][l]{$|\Y|=2$}
\psfrag{B }[l][l]{$|\Y|=3$}
\psfrag{C }[][]{$|\Y|=4$}
\psfrag{-5}[r][r]{$10^{-5}$}
\psfrag{-10}[r][r]{$10^{-10}$}
\psfrag{-15}[r][r]{$10^{-15}$}
\psfrag{-20}[r][r]{$10^{-20}$}
\psfrag{-25}[r][r]{$10^{-25}$}
\psfrag{-30}[r][r]{$10^{-30}$}
\psfrag{-35}[r][r]{$10^{-35}$}
\psfrag{-40}[r][r]{$10^{-40}$}
\psfrag{-45}[r][r]{$10^{-45}$}
\psfrag{-50}[r][r]{$10^{-50}$}
\psfrag{0}[r][l]{1}
\psfrag{1}[c][c]{1}
\psfrag{2}[c][c]{2}
\psfrag{3}[c][c]{3}
\psfrag{4}[ct][ct]{$\underset{\displaystyle |\X|}{4}$}
\psfrag{5}[c][c]{5}
\psfrag{6}[c][c]{6}
\psfrag{7}[c][c]{7}
\begin{center}
\epsfig{file=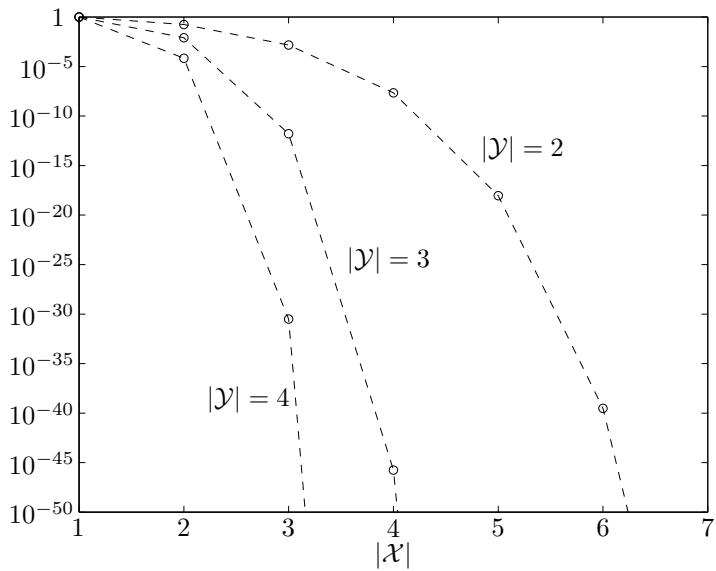,width =.75\textwidth}
\end{center}
\caption{\label{plot:fig}
The ordinate gives the fraction of subsets closed under permutation
on logarithmic scale
given the cardinality of the search space $\X$.
The different curves correspond to different cardinalities
of the codomain $\Y$.}
\end{figure}
Figure \ref{plot:fig} shows a plot of the fraction of non-empty
subsets \clup{}, i.e.,
\begin{equation*}
\left({2^{\binom{|\X|+|\Y|-1}{|\X|}} -1}\right)\Big/\left({2^{\left(|\Y|^{|\X|}\right)}-1}\right)
\enspace,
\end{equation*}
versus the cardinality of $\X$ for different values of $|\Y|$.
The fraction decreases for increasing $|\X|$ as well as  for
increasing $|\Y|$. Already for small $|\X|$ and $|\Y|$
the fraction almost vanishes, e.g., for a Boolean
function $f:\{0,1\}^3\to\{0,1\}$ the fraction is $\ll 10^{-170}$.

%%%%%%%%%%%%%%%%%%%%%%%%%%%%%%%%%%%%%%%%%%%%%%%%%%%%%%%%%%%%%%%%%%%%%%%%%%%
\section{Search Spaces with Neighborhood Relations}\label{toppology:sec}
%%%%%%%%%%%%%%%%%%%%%%%%%%%%%%%%%%%%%%%%%%%%%%%%%%%%%%%%%%%%%%%%%%%%%%%%%%%
In the previous section, we have shown that the fraction of subsets
\clup{} is close to zero already for small search and cost-value
spaces. Still, the absolute number of subsets \clup{} grows rapidly
with increasing $|\X|$ and $|\Y|$.  What if these classes of functions
are the ``important'' ones, i.e., those we are dealing with in
practice?  In this section, we define some quite general constraints
on functions important in
practice that induce classes of functions that are not \clup{}
%Our consideration are similar to the ones in \cite{whitley:99}.

We believe that two assumptions can be made for most of the functions
we are dealing with in real-world optimization: First, the search
space has some structure.  Second, the set of objective functions we
are interested in fulfills some constraints defined based on this
structure.  More formally, there exists a non-trivial neighborhood
relation on $\X$ based on which constraints on the set of functions
under consideration are formulated.  For example, with respect to a
neighborhood relation we can define concepts like ruggedness or local
optimality and constraints like upper bounds on the ruggedness or on
the maximum number of local minima.  Intuitively, it is likely that in
a function class \clup{} there exists a function that violates such
constraints.

We define a simple neighborhood relation on $\X$ as a symmetric
function $n:\X\times\X\to\{0,1\}$.  Two elements $x_i,x_j\in\X$ are
called neighbors iff $n(x_i,x_j)=1$.  We call a neighborhood
non-trivial iff $\exists x_i,x_j\in\X: x_i\neq x_j \,\wedge\,
n(x_i,x_j)=1$ and $\exists x_k,x_l\in\X: x_k\neq x_l \,\wedge\,
n(x_k,x_l)=0$.  It holds:
\begin{theorem}
A non-trivial neighborhood on $\X$ is not invariant under permutations
of $\X$.
\end{theorem}
\begin{proof}
It holds $\exists x_i,x_j,x_k,x_l\in\X:\,
x_i\neq x_j \,\wedge\, x_k\neq x_l\,\wedge\,
n(x_i,x_j)=0\,\wedge\, n(x_k,x_l)=1$.
For any permutation $\pi$ that maps $x_i$ and $x_j$ onto
$x_k$ and $x_l$, respectively, the invariance property,
$\forall a,b\in\X:\,n(x_a,x_b)=n(\pi(x_a), \pi(x_b))$,
is violated.%\qed
\end{proof}
\begin{remark}
  Assume the search space $\X$ can be decomposed as $\X=\X_1 \times
  \dots\times \X_l, l>1$ and let on one component $\X_i$ exist a
  non-trivial neighborhood $n_i:\X_i\times\X_i\to\{0,1\}$. This
  neighborhood induces a non-trivial neighborhood on $\X$, where two
  points are neighbored iff their $i$-th components are neighbored
  with respect to $n_i$. Thus, the constraints discussed below need
  only refer to a single component.
\end{remark}
\begin{remark}
  The neighborhood relation need not be the canonical one (e.g.,
  Ham\-ming-distance for Boolean search spaces).  Instead, it can be
  based on ``phenotypic'' properties (e.g., if integers are encoded by
  bit-strings, then the bit-strings can be defined as neighbored iff
  the corresponding integers are).
\end{remark}

Now we describe some constrains that are defined with respect to a
neighborhood relation and are---to our minds---relevant in
practice. For this purpose, we assume a metric
$\m_{\Y}:\Y\times\Y\rightarrow\mathbbm R$ on $\Y$, e.g., in the
typical case of real-valued fitness function $\Y\subset\mathbbm R$ the
Euclidean distance.

First, we show how a constraint on steepness (closely related
to the concept of \emph{strong causality})
leads to a set of functions that is not \clup{}
Based on a neighborhood relation on the search space, we can define a
simple measure of maximum steepness of a function $f\in\F$ by
\[
s^{\max}(f)=\max_{x_i,x_j\in\X\,\wedge\, n(x_i,x_j) = 1 }  \m_{\Y}(f(x_i),f(x_j))
\enspace.
\]
Further, for a function $f\in F$, we define the diameter of its range as
\[
d^{\max}(f)=\max_{x_i,x_j\in\X}  \m_{\Y}(f(x_i),f(x_j))
\enspace.
\]
\begin{corollary}
  If the maximum steepness $s^{\max}(f)$ of every function $f$ in a
  non-empty subset $F\subset \F$ is constrained to be smaller than the
  maximal possible $\max_{f\in F} d^{\max}(f)$, then $F$ is not \clup
\end{corollary}
\begin{proof}
  Let $g=\text{arg}\max_{f \in F} d^{\max}(f)$ and let $x_i$ and $x_j$
  be two points with property $\m(g(x_i),g(x_j))=d^{\max}(g)$. Since
  the neighborhood on $\X$ is non-trivial there exist two neighboring
  points $x_k$ and $x_l$. There exists a permutation $\pi$ that maps
  $x_i$ and $x_j$ on $x_k$ and $x_l$.  If $F$ is \clup{}, the function
  $g\circ \pi$ is in $F$. This function has steepness
  $s^{\max}(g\circ\pi)=d^{\max}(g)=\max_{f\in F} d^{\max}(f)$, which
  contradicts the steepness-constraint.%\qed
\end{proof}

As a second constraint, we consider the number of local minima, which
is often regarded as a measure of complexity \cite{whitley:99}.  For a
function $f \in \F$ a point $x \in \X$ is a local minimum iff
$f(x)<f(x_i)$ for all neighbors $x_i$ of $x$.  Given a function $f$
and a neighborhood relation on $\X$, we define $l^{\max}(f)$ as the maximal
number of minima that functions with the same $\Y$-histogram as $f$
can have (i.e., functions where the number of $\X$-values that are
mapped to a certain $\Y$-value are the same as for $f$, see
appendix). In the appendix we prove that for any two functions $f,g$
with the same $\Y$-histogram there exists a permutation
$\pi\in\pot(\X)$ with $f\circ\pi = g$. Thus, it follows:
\begin{corollary}
  If the number of local minima of every function $f$ in a non-empty
  subset $F\subset \F$ is constrained to be smaller than the maximal
  possible $\max_{f\in F} l^{\max}(f)$, then $F$ is not \clup
\end{corollary}
\noindent For example, consider pseudo-Boolean function $\{0,1\}^n\to{\mathbbm
R}$ and let two points be neighbored iff they have Hamming-distance one.
Then the maximum number of local minima is  $2^{n-1}$.

%%%%%%%%%%%%%%%%%%%%%%%%%%%%%%%%%%%%%%%%%%%%%%%%%%%%%%%%%%%%%%%%%%%%%%%%%%%
\section{Conclusion}
%%%%%%%%%%%%%%%%%%%%%%%%%%%%%%%%%%%%%%%%%%%%%%%%%%%%%%%%%%%%%%%%%%%%%%%%%%%
Based on the results in \cite{schuhmacher:01}, we have shown that the
statement ``I'm only interested in a subset $F$ of all possible
functions, so the NFL theorems do not apply'' is true with a
probability close to one (if $F$ is chosen uniformly and 
$\Y$ and $\X$ have reasonable cardinalities).
%$|Y|>1$ and $|X|>6$). 
Further, the statements ``In my application domain,
functions with maximum number of local minima are not realistic'' and
``For some components, the objective functions under consideration
will not have the maximal possible steepness'' lead to scenarios where
NFL does not hold.

%%%%%%%%%%%%%%%%%%%%%%%%%%%%%%%%%%%%%%%%%%%%%%%%%%%%%%%%%%%%%%%%%%%%%%%%%%%
\subsection*{Acknowledgments}
%%%%%%%%%%%%%%%%%%%%%%%%%%%%%%%%%%%%%%%%%%%%%%%%%%%%%%%%%%%%%%%%%%%%%%%%%%%
We thank Hannes Edelbrunner for fruitful discussions and
Thomas Jansen, Stefan Wiegand, and Michael H\"usken for their
comments on the manuscript.
This work was supported by the
DFG, grant Solesys, number SE251/41-1.

%%%%%%%%%%%%%%%%%%%%%%%%%%%%%%%%%%%%%%%%%%%%%%%%%%%%%%%%%%%%%%%%%%%%%%%%%%%
\appendix
\section{Proof of Theorem \ref{main:thm}}
%%%%%%%%%%%%%%%%%%%%%%%%%%%%%%%%%%%%%%%%%%%%%%%%%%%%%%%%%%%%%%%%%%%%%%%%%%%
For the proof, we use  the concepts of $\Y$-histograms:
%\begin{definition}
  We define a \emph{\Y-histogram} (\emph{histogram} for short) as a
  mapping $h:\, \Y \to \NNN_0$ such that $\sum_{y \in \Y} h(y) =
  |\X|$.  The set of all histograms is denoted $\H$. With any function
  $f:\, \X \to \Y$ we associate the histogram $h(y)=| f^{-1}(y)|$ that
  counts the number of elements in $\X$ that are mapped to the same
  value $y \in \Y$ by $f$.  Herein, $f^{-1}(y), y\in\Y$ returns the
  preimage $\{x|f(x) = y\}$ of $f$.  Further, we call two functions
  $f,g$ \emph{$h$-equivalent} iff they have the same histogram and we
  call the corresponding $h$-equivalence class
  $B_h \subseteq\F$ containing all function with histogram $h$ a \emph{basis class}.
%\end{definition}
 Before we prove theorem
\ref{main:thm}, we consider the following lemma that
gives some basic properties of basis classes.
\begin{lemma}\label{lemma:lem}
  \begin{enumerate}
  \item\label{a:lem} There exist \[\binom{|\X|+|\Y|-1}{|\X|}\]pairwise disjoint
    basis classes and \[\bigcup_{h\in\H} B_h = \F\enspace.\]
  \item\label{b:lem} Two functions $f,g\in\F$ are $h$-equivalent iff
    there
    exists a permutation $\pi$ of $\X$ such that $f \circ \pi = g$.
  \item\label{c:lem} $B_h$ is equal to the permutation orbit of any function $f$
    with histogram $h$, i.e.,
    \begin{equation*}
      B_h = \bigcup_{\pi\in\pot(\X)} \{ f \circ \pi \} \enspace.
    \end{equation*}
  \item\label{d:lem} Any subset $F\subseteq \F$ that is c.u.p. is uniquely defined by
    a union of pairwise disjoint basis classes.
  \end{enumerate}
\end{lemma}
\begin{proof}
\begin{enumerate}
\item
The number $|\H|$ of different histograms is given by
\begin{equation*}
\binom{|\X|+|\Y|-1}{|\X|} \enspace,
\end{equation*}
%see, e.g., \cite{feller:68}.
i.e., the number of \emph{distinguishable distributions}
(e.g., \cite{feller:68}, p.\ 38).
Two basis classes $B_{h_1}$ and $B_{h_2}$, $h_1\neq h_2$,
are disjoint because functions in different
basis classes have different histograms. The union
$\bigcup_{h\in\H} B_h = \F$ because every function in $\F$ has a
histogram.%and $\F$ is \clup.

\item
Let $f,g\in\X$ be two functions with same histogram $h$. Then,
for any $y \in \Y$, $f^{-1}(y)$ and $g^{-1}(y)$
are equal in size and there exists a bijective function 
$\pi_y$
between these
two subsets. Then the bijection 
\begin{equation*}
\pi(x)=\pi_y(x)\quad\text{, where } y=f(x)\enspace,
%\text{ if } x\in f^{-1}(y)
\end{equation*}
defines a unique permutation such that  $f\circ \pi=g$. 
Thus, $h$-equivalence implies existence of a permutation.
On the other hand,
the histogram of a function is invariant under
permutation since for any $y\in\Y$ and $\pi\in\pot(\X)$
\begin{equation*}
\big|(f\circ\pi)^{-1}(y)\big|=
\sum_{x\in\X}\delta(y,f(\pi(x)))=
\sum_{x\in\X}\delta(y,f(x))=
% = \big|\pi[f^{-1}(y)]\big| =
\big|f^{-1}(y)\big| \enspace,
\end{equation*}
because $\pi$ is bijective and the addends can be resorted.
Thus, existence of a permutation implies $h$-equivalence.

\item For a function $f$ with histogram $h$, let
$O_f=\bigcup_{\pi\in\pot(\X)} \{ f \circ \pi \}$ be the orbit of $f$
under permutations $\pi$.  By (\ref{b:lem}), all functions in $O_f$ have
the same histogram and thus $O_f \subseteq B_h$. On the other hand,
for any functions $g \in B_h$ there exists by (\ref{b:lem}) a
permutation $\pi$ such that $f \circ \pi = g$ and thus $B_h \subseteq
O_f$.

\item For a subset $F \subseteq \F$, let $F_h = B_h \cap F$ (i.e.,
$F_h$ contains all functions in $F$ with the same histogram $h$).  By
(\ref{a:lem}), all $F_h$ are pairwise disjoint and $F=\bigcup_{h\in\H}
F_h$.  Suppose $F_h\neq \emptyset$: Since $F$ is \clup{}
there exists a function $f \in F_h$ that spans the orbit $B_h$. Thus
$B_h\subseteq F$ and therefore $F_h=B_h$.  Because basis classes are
disjoint, the union
\begin{equation*}
F=\bigcup_{h:\,h\in\H \,\wedge\, F_h \neq \emptyset} B_h
%\enspace.
\end{equation*}
is unique.
\end{enumerate}%\qed
\end{proof}

\begin{proof}[Proof of theorem \ref{main:thm}]
  By lemma \ref{lemma:lem}(\ref{a:lem}), the number of different basis
  classes is given by
\begin{equation*}
%\binom{(|\X|+1) + (|\Y|-1) - 1}{|\Y|-1} =
%\binom{|\X|+|\Y|-1}{|\Y|-1}
\binom{|\X|+|\Y|-1}{|\X|}
\enspace.
  \end{equation*}
  The number of different, non-empty unions of basis classes (equal to
  the cardinality of power set of the set of all basis classes minus
  one for the empty set) is given by
\begin{equation*}
2^{\binom{|\X|+|\Y|-1}{|\X|}}-1
\enspace.
\end{equation*}
By lemma \ref{lemma:lem}(\ref{d:lem}), this is the number of non-empty
subsets of $\F$
that are \clup%\qed
\end{proof}

{ \small
\bibliographystyle{abbrv} \bibliography{nfl} }

\end{document}